\documentclass[runningheads]{llncs}
\usepackage[T1]{fontenc}
\usepackage{graphicx}
\usepackage{bm}
\usepackage{cite}
\usepackage{tikz}
\usepackage{pgfplots}
\usepackage{pgfplotstable}
\usepackage{booktabs}
\usepackage[final]{microtype}
\usepackage{subfigure}

\usepackage{tikz}
\usetikzlibrary{positioning}
\usepackage{amssymb}
\usepackage{amsfonts}
\usepackage{amsmath}
\usepackage{xifthen}
\usepackage{xparse}
\usepackage{dsfont}
\usepackage{mathtools}
\usepackage{gensymb}

\newcommand{\R}{\mathbb R}

\newcommand{\G}{G}
\newcommand{\XX}{\mathcal{X}}
\newcommand{\YY}{\mathcal{Y}}

\usepackage{import}
\usepackage{pgfplots}
\usepackage{tikz-cd}

\graphicspath{ {figures/} }

\usepackage{adjustbox}
\usepackage{hhline}

\usepackage{color}

\usepackage{hyperref}
\renewcommand{\d}{{\,\mathrm{d}}}

\usepackage{rotating}
\definecolor{plotblue}{HTML}{41658A}
\definecolor{plotred}{HTML}{E4572E}
\definecolor{plotyellow}{HTML}{17BEBB}
\definecolor{plotgreen}{HTML}{248232}

\begin{document}
\title{Scale-Equivariant Deep Learning for 3D Data}

\author{Thomas Wimmer\inst{1} \and
Vladimir Golkov\inst{1,2} \and
Hoai Nam Dang\inst{3} \and
Moritz Zaiss\inst{3} \and
Andreas Maier\inst{3} \and
Daniel Cremers\inst{1,2}
}
\authorrunning{T. Wimmer et al.}
\institute{%Computer Vision Group,\\School of Computation, Information and Technology,\\
Technical University of Munich\and
Munich Center for Machine Learning\and
%Institute of Neuroradiology, University Clinic of Erlangen,\\
Friedrich-Alexander University Erlangen-Nuremberg\\
%\and
%High‐field Magnetic Resonance Center,\\Max Planck Institute for Biological Cybernetics, Tübingen\and
%Pattern Recognition Lab, Friedrich-Alexander University Erlangen-Nuremberg
\email{thomas.m.wimmer@tum.de}, \email{vladimir.golkov@tum.de}}

\maketitle
\begin{abstract}
The ability of convolutional neural networks (CNNs) to recognize objects regardless of their position in the image is due to the translation-equivariance of the convolutional operation. Group-equivariant CNNs transfer this equivariance to other transformations of the input. Dealing appropriately with objects and object parts of different scale is challenging, and scale can vary for multiple reasons such as the underlying object size or the resolution of the imaging modality.
In this paper, we propose a scale-equivariant convolutional network layer for three-dimensional data that guarantees scale-equivariance in 3D CNNs.
Scale-equivariance lifts the burden of having to learn each possible scale separately, allowing the neural network to focus on higher-level learning goals, which leads to better results and better data-efficiency.
We provide an overview of the theoretical foundations and scientific work on scale-equivariant neural networks in the two-dimensional domain. We then transfer the concepts from 2D to the three-dimensional space and create a scale-equivariant convolutional layer for 3D data.
Using the proposed scale-equivariant layer, we create a scale-equivariant {U-Net} for medical image segmentation and compare it with a non-scale-equivariant baseline method. Our experiments demonstrate the effectiveness of the proposed method in achieving scale-equivariance for 3D medical image analysis.\footnote{We publish our code using PyTorch for further research and application: 

\url{https://github.com/wimmerth/scale-equivariant-3d-convnet}}

\keywords{Scale-Equivariance  \and Data Efficiency \and Segmentation}
\end{abstract}

\section{Introduction}\label{sec:introduction}
One of the greatest advantages of convolutional neural networks (CNNs) is their \emph{equivariance} under spatial shifts (translations), i.e.\ a mathematically guaranteed ability to recognise objects and object parts at any positions they might appear at in images.
Recent methods additionally provide equivariance under other transformations of the input such as rotation or scaling, in order to guarantee that features get detected well at different orientations or sizes.
Scale-equivariance, i.e.~guaranteed detection of features across various scales/sizes, is a beneficial property of neural networks because an object or object part can have different sizes in different images, and because combining datasets with different image resolutions, for example in medical imaging, allows for constructing richer, more informative training datasets than if large parts of the data are left out.
In practice, scale-equivariance often leads to better results~\cite{sosnovik2019scale, sosnovik2021disco, zhu2019scale}.

In deep learning, data augmentation is often used as a means to improve the recognition of scaled features. However, data augmentation does not guarantee equivariance, it merely tries to approximate it, and imposes an additional burden on the neural network to learn each possible scale of each feature separately. In many cases, results of such learned equivariance are worse than with guaranteed equivariance~\cite{cohen2016group, bekkers2019b}.

Therefore, intensive research has been conducted in recent years on the development of scale-equivariant neural networks~\cite{sosnovik2019scale, lindeberg2021scale}. So far, these methods have been limited to the two-dimensional case.
Since the detection of features of different scales is an important aspect in 3D machine learning tasks as well, in the present work we extend the concept of scale-equivariance to three dimensions.
We propose novel scale-equivariant neural network layers for 3D data, including convolutions, normalization and pooling, that can be used instead of usual 3D neural network layers to achieve scale-equivariance. Medical image analysis, such as brain tumor segmentation, has been shown to benefit from aggressive data augmentation to approximate scale-equivariance~\cite{isensee2020nnu} and as 3D data in many applications are available at various image resolutions, it is of high relevance to extend scale-equivariance to the 3D setting.

Scale-equivariant neural networks have been shown to outperform other neural networks especially in the low data regime \cite{zhu2019scale, sosnovik2021disco}, which is particularly interesting for 3D applications such as MRI and datasets of rare diseases, as there is usually much less training data available than for example photographs in the two-dimensional case.
In this paper, we first present the theoretical foundations for scale-equivariant convolutions and review previous works in this field. We lift the concept of scale-equivariant convolutions to the three-dimensional space and evaluate the performance of the proposed layers through a series of experiments based on the brain tumor segmentation task. To demonstrate the efficacy of our approach, we evaluate it against a baseline method based on standard convolutions which are only translation-equivariant.

\section{Related Work}\label{sec:relatedwork}

When image features can appear at a variety of sizes and locations, a primitive approach
is to train a neural network that is neither equivariant nor is specifically designed to easily approximate equivariance.
Instead, primitive methods tediously learn to approximate equivariance, and their training dataset must contain differently transformed features. That dataset might be obtained by data augmentation (random transformations) if the original dataset lacks such diversity.
The additional burden of learning every feature at every possible size distracts the training from the main learning goals, and in practice yields suboptimal results.

Better results can be achieved by methods that are designed to slightly facilitate the learning of approximate equivariance. An example is to use a branch of a neural network that decides how to transform (for example scale) the input before passing it to another neural-network branch, thus facilitating (but not guaranteeing) equivariance only for features that scale jointly but not independently\cite{girshick2014rich,jaderberg2015spatial}. Another example are capsule networks, which encourage the separation of visual features and their poses (for example scale) in the latent representations by computing deeper features in ways that benefit from such a separation\cite{sabour2017dynamic}.
Other examples are scale-dependent pooling of latent features~\cite{yang2016exploit}, or the addition of downsampling and/or upsampling branches in the network~\cite{chen2019drop}.

The best results are usually obtained by methods that mathematically guarantee equivariance.
Examples include translation-equivariance that is achieved by using convolutional networks, equivariance under 2D rotations and translations\cite{cohen2016group,cohen2016steerable,weiler2018learning}, 3D rotations and translations~\cite{thomas2018tensor,weiler20183d,muller2021rotation,liu2022group}, or 2D scalings and translations~\cite{kanazawa2014locally,xu2014scale,marcos2018scale,worrall2019deep,sosnovik2021disco,ghosh2019scale,naderi2020scale,sosnovik2019scale,zhu2019scale,lindeberg2021scale,jansson2021scale}.
Special cases of scale-equivariant neural networks use invariant rather than more generally equivariant layers not only as the last layer but also throughout~\cite{kanazawa2014locally, ghosh2019scale}, which does not allow information about the relative scale of different features to be used to compute deeper features. Another special case keeps information about different scales separated until the last network layer~\cite{xu2014scale, lindeberg2021scale, jansson2021scale}, with a similar effect.

Neural networks achieve scale-equivariance by using differently scaled versions of convolutional filters (or, equivalently, of the feature maps). Recent works reduce discretization artifacts during scaling by representing filters using certain truncated bases, for example Hermite polynomials with Gaussian envelopes~\cite{sosnovik2019scale, zhu2019scale}, Gaussian derivatives~\cite{lindeberg2021scale, jansson2021scale}, or radial harmonics~\cite{ghosh2019scale, naderi2020scale}.
In the medical domain, scale-equivariant neural networks have so far successfully been applied for histopathology image segmentation~\cite{yang2022scale} and 2D MRI reconstruction~\cite{gunel2022scale}.

The present paper extends the theory of scale-equivariant neural networks to~3D. We base our method on the work by \cite{sosnovik2019scale} for the 2D case as it achieved state-of-the-art performance while being the most flexible approach.

\section{Methods}\label{sec:methods}
In this section, we present the theory behind scale-equivariant convolutions and use them to propose a novel scale-equivariant layer for 3D data. We furthermore propose a novel scale-equivariant 3D~U-Net that can be used in tasks like medical image segmentation.

A \emph{group} $(\G, \cdot)$ is defined as a set $\G$ closed under an associative binary operation $\cdot: \G \times \G \rightarrow \G$, with an identity element $e \in \G$, and where every element has an inverse also in the set.
A group~$(\G, \cdot)$ is often simply denoted by~$\G$, with the binary operation~$\cdot$ implied.
A mapping $\Phi: \XX \rightarrow \YY$ is \emph{equivariant} under actions of the group $\G$ when
\begin{equation}
    \Phi(L_g[f]) = L'_g[\Phi(f)] \quad \forall f \in \XX \quad \forall g \in \G,
\end{equation}
where~$f$ is the input of $\Phi$, for example a medical image or latent feature map, and~$L_g$ and $L'_g$ are the actions of $g \in \G$ on $\XX$ and $\YY$, respectively.
When $L'_g$ is the identity for all $g\in \G$, then $\Phi$ is called invariant under actions of the group $\G$.

The \emph{scaling group} can be defined as $H = (\R_{>0},\cdot)$, i.e.~consisting of positive scaling factors and with multiplication $\cdot$ as the binary operation.
A scale transformation $L_s$ of a $d$-dimensional real-valued image~${f: \R^d\rightarrow \R}$ (for example a 3D MRI volume with $d=3$) by a scaling factor~$s\in H$ (i.e.~${s\in \R_{>0}}$) can be formalized as a scale transformation of image coordinates~$x \in \R^d$ and can then be formulated as ${[L_s[f]](x) = f(s^{-1}x)}$. We refer to the transformation as upscale if~$s > 1$ and downscale if~$s < 1$.

We want our neural network to be equivariant under scalings and translations. By combining the scaling group~$H$ with the group~$T$ of translations using the semi-direct product, we obtain the \emph{group of scalings and translations} $HT = \{(s,t) \mid s \in H, \; t \in T\}$ with $s \in \R_{>0}, \; t \in \R^d$, sometimes denoted as $HT \cong \R_{>0} \ltimes \R^d$~\cite{worrall2019deep}, with the identity element~$(1,0)$, the binary operation
\begin{equation}
    (s_2,t_2) \cdot (s_1, t_1) = (s_2s_1,s_2t_1 + t_2),
\end{equation}
and the inverse $(s,t)^{-1} = (s^{-1}, -s^{-1}t)$.
As can be seen in this definition, the order of applying scaling and translation matters, i.e.~$(s,t)=(s,0)\cdot(1,t)\neq(1,t)\cdot(s,0)$).

The group actions $L_{(s,t)}$ of $(s,t) \in HT$ on functions~$f : \R^d \rightarrow \R^C$ (input images) and on functions~$h : HT \rightarrow R^C$ (latent feature maps) are defined as follows:
\begin{equation}\label{eq:HT-group-actions}
\begin{split}
    L_{(s,t)}[f](x) &= f(s^{-1}(x - t)),\\
    L_{(s,t)}[h](s', t') &= h((s,t)^{-1}(s',t')) = h(s^{-1}s', s^{-1}(t' - t)).
\end{split}
\end{equation}

A \emph{group-convolution}~$\star_\G$~\cite{cohen2016group, sosnovik2019scale, cohen2021equivariant}, using a locally compact group $G$, of a function $f : X \rightarrow \R^C$, for example a medical image (e.g.~${X = \mathbb{Z}^3}$) or a latent feature map (${X = \G}$), with a function ${\psi : X \rightarrow \R^C}$ (e.g.~a convolutional filter), where~$C\in\mathbb{N}$ is referred to as the number of \emph{channels} of the input~$f$, is defined as
\begin{equation} \label{eq:g-convolution}
    [f\star_\G\psi](g) = \int_{x \in X} f(x) [L_g [\psi]](x) \d\mu(x),
\end{equation}
where $g \in \G$ with its according group action $L_g$, and $\mu(x)$ is the Haar measure.
Thus, the output of a group convolution is a feature map defined on the group $\G$.
When that feature map is used as an input to a subsequent group convolution, the filters must also be defined on $\G$, because the input and filter in a group convolution are defined on the same space.

Group-convolutional network layers generalize group convolutions from having several input channels (which we included in the definition of group convolutions above) to additionally having several output channels. This is achieved by performing a group convolution of the input with each of several filters, each yielding a separate output channel.
Several channels are used in neural networks to disentangle different features.
The filters in these layers are represented as a weighted sum of fixed (truncated-)basis functions. The weights in that sum are the trainable parameters of the layer.
 
Note that when the group $\G$ is the group $T = (\R^d,+)$ of translations, with addition $+$ as the binary operation, and the group action $L_t$ of a translation $t\in T$ is $L_t[f](x) = f(x-t)$, then Eq.~\eqref{eq:g-convolution} is the well-known standard convolution.

Group convolution using the (continuous) scaling group computes a value for each of the infinitely many elements of the group. In order to make the computational and output memory requirements finite, usually a discrete subgroup of the continuous group is used, and the subgroup is additionally truncated to a semigroup if it is still infinite~\cite{bekkers2019b,sosnovik2019scale}. \label{sec:truncated-group}
A discrete subgroup of the scaling group can be constructed as ${\{..., s^{-1}, 1, s^1, s^2, ...\}}$ with a base scale $s$, for example $s=0.9$, and truncated for example to $\{1,s,s^2,s^3\}$.
The truncation breaks the equivariance under scales beyond the truncation boundary, but is still locally correct, i.e.~guarantees equivariance under discrete scales within the truncated discretized group of scales~\cite{worrall2019deep}.

\begin{figure}[bt]
    \centering
    \subfigure{
\centering
\begin{tikzpicture}[scale = 0.55]
  \begin{axis}[
    axis lines = middle,
    xmin = -6.2, xmax = 6.2,
    ymin = -6.2, ymax = 6.2,
    domain = -6 : 6,
    xlabel = $x$,
    ylabel = $\psi(x)$,
    samples = 100,
    thick,
    legend style = {at = {(0.6, 0.2)}, anchor = west, draw = none, },
    legend cell align={left},
    declare function = {
      H0(\x) = 1 * exp(- abs(\x)^2 / 2);
      H1(\x) = 2 * \x * exp(- abs(\x)^2 / 2);
      H2(\x) = (4 * \x^2 - 2) * exp(- abs(\x)^2 / 2);
      H3(\x) = (8 * \x^3 - 12 * \x) * exp(- abs(\x)^2 / 2);
      H0s(\x) = 1  * exp(- abs(\x)^2 / (2 * 2^2)) / 2;
      H1s(\x) = \x * exp(- abs(\x)^2 / (2 * 2^2)) / 2;
      H2s(\x) = (4 * (\x / 2)^2 - 2) * exp(- abs(\x)^2 / (2 * 2^2)) / 2;
      H3s(\x) = (8 * (\x / 2)^3 - 12 * \x / 2) * exp(- abs(\x)^2 / (2 * 2^2)) / 2;
    }, ]
    \addplot[plotred] {H0(x)}; \addlegendentry{$n=0, \sigma=1.0$};
    \addplot[plotred, dashed] {H0s(x)}; \addlegendentry{$n=0, \sigma=2.0$};
  \end{axis}
\end{tikzpicture}
}
\subfigure{
\centering
\begin{tikzpicture}[scale = 0.55]
  \begin{axis}[
    axis lines = middle,
    xmin = -6.2, xmax = 6.2,
    ymin = -6.2, ymax = 6.2,
    domain = -6 : 6,
    xlabel = $x$,
    ylabel = $\psi(x)$,
    samples = 100,
    thick,
    legend style = {at = {(0.6, 0.2)}, anchor = west, draw = none, },
    legend cell align={left},
    declare function = {
      H0(\x) = 1 * exp(- abs(\x)^2 / 2);
      H1(\x) = 2 * \x * exp(- abs(\x)^2 / 2);
      H2(\x) = (4 * \x^2 - 2) * exp(- abs(\x)^2 / 2);
      H3(\x) = (8 * \x^3 - 12 * \x) * exp(- abs(\x)^2 / 2);
      H0s(\x) = 1  * exp(- abs(\x)^2 / (2 * 2^2)) / 2;
      H1s(\x) = \x * exp(- abs(\x)^2 / (2 * 2^2)) / 2;
      H2s(\x) = (4 * (\x / 2)^2 - 2) * exp(- abs(\x)^2 / (2 * 2^2)) / 2;
      H3s(\x) = (8 * (\x / 2)^3 - 12 * \x / 2) * exp(- abs(\x)^2 / (2 * 2^2)) / 2;
    }, ]
    \addplot[plotblue] {H1(x)}; \addlegendentry{$n=1, \sigma=1.0$};
    \addplot[plotblue, dashed] {H1s(x)}; \addlegendentry{$n=1, \sigma=2.0$};
  \end{axis}
\end{tikzpicture}
}
\subfigure{
\centering
\begin{tikzpicture}[scale = 0.55]
  \begin{axis}[
    axis lines = middle,
    xmin = -6.2, xmax = 6.2,
    ymin = -6.2, ymax = 6.2,
    domain = -6 : 6,
    xlabel = $x$,
    ylabel = $\psi(x)$,
    samples = 100,
    thick,
    legend style = {at = {(0.6, 0.2)}, anchor = west, draw = none, },
    legend cell align={left},
    declare function = {
      H0(\x) = 1 * exp(- abs(\x)^2 / 2);
      H1(\x) = 2 * \x * exp(- abs(\x)^2 / 2);
      H2(\x) = (4 * \x^2 - 2) * exp(- abs(\x)^2 / 2);
      H3(\x) = (8 * \x^3 - 12 * \x) * exp(- abs(\x)^2 / 2);
      H0s(\x) = 1  * exp(- abs(\x)^2 / (2 * 2^2)) / 2;
      H1s(\x) = \x * exp(- abs(\x)^2 / (2 * 2^2)) / 2;
      H2s(\x) = (4 * (\x / 2)^2 - 2) * exp(- abs(\x)^2 / (2 * 2^2)) / 2;
      H3s(\x) = (8 * (\x / 2)^3 - 12 * \x / 2) * exp(- abs(\x)^2 / (2 * 2^2)) / 2;
    }, ]
    \addplot[plotgreen] {H2(x)}; \addlegendentry{$n=2, \sigma=1.0$};
    \addplot[plotgreen, dashed] {H2s(x)}; \addlegendentry{$n=2, \sigma=2.0$};
  \end{axis}
\end{tikzpicture}
}
\subfigure{
\centering
\begin{tikzpicture}[scale = 0.55]
  \begin{axis}[
    axis lines = middle,
    xmin = -6.2, xmax = 6.2,
    ymin = -6.2, ymax = 6.2,
    domain = -6 : 6,
    xlabel = $x$,
    ylabel = $\psi(x)$,
    samples = 100,
    thick,
    legend style = {at = {(0.6, 0.2)}, anchor = west, draw = none, },
    legend cell align={left},
    declare function = {
      H0(\x) = 1 * exp(- abs(\x)^2 / 2);
      H1(\x) = 2 * \x * exp(- abs(\x)^2 / 2);
      H2(\x) = (4 * \x^2 - 2) * exp(- abs(\x)^2 / 2);
      H3(\x) = (8 * \x^3 - 12 * \x) * exp(- abs(\x)^2 / 2);
      H0s(\x) = 1  * exp(- abs(\x)^2 / (2 * 2^2)) / 2;
      H1s(\x) = \x * exp(- abs(\x)^2 / (2 * 2^2)) / 2;
      H2s(\x) = (4 * (\x / 2)^2 - 2) * exp(- abs(\x)^2 / (2 * 2^2)) / 2;
      H3s(\x) = (8 * (\x / 2)^3 - 12 * \x / 2) * exp(- abs(\x)^2 / (2 * 2^2)) / 2;
    }, ]
    \addplot[plotyellow] {H3(x)}; \addlegendentry{$n=3, \sigma=1$};
    \addplot[plotyellow, dashed] {H3s(x)};\addlegendentry{$n=3, \sigma=2$};
  \end{axis}
\end{tikzpicture}
}
    \caption{Basis functions defined as Hermite polynomials $H_n$ (of different degrees~$n$) with Gaussian envelopes, scaled with different scales~$\sigma$.
    The kernel basis for three-dimensional scale-equivariant convolutions is formed from the multiplication of three oriented basis functions (oriented in the $x$, $y$, and $z$ directions) with equal or different degrees of Hermite polynomials.}
    \label{fig:basis}
\end{figure}
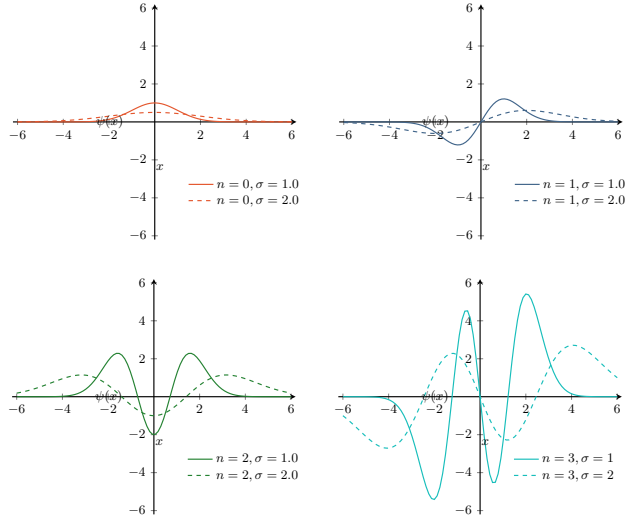

We propose scale-equivariant 3D convolutional layers by performing group-convolutions using the discretized truncated version of the group.
For filters, truncated bases based on Hermite polynomials with a Gaussian envelope (see Fig.~\ref{fig:basis}) were shown to work the best in practice for scale-equivariant deep learning due to reduced interpolation artifacts during scaling~\cite{sosnovik2019scale}. We generalize the construction of such bases to 3D. We multiply combinations of 1D Hermite polynomials of increasing order (with the maximum order being a hyperparameter) along each of the three dimensions and apply 3D Gaussian envelopes. \label{sec:basis}

We used replicate-padding along the ``scale dimension'' of feature maps on the group because this technique is a good compromise between the truncation of the group and the imprecision of equivariance due to padding~\cite{zhu2019scale}.
We also create additional scale-equivariant layers, namely the addition of bias terms, pointwise nonlinearities, max- and average-pooling over subgroups (e.g.\ the group of scalings), and normalization (batch normalization, instance normalization) for the 3D setting, analogously to the 2D setting~\cite{sosnovik2019scale}. Neural networks consisting of combinations of these layers are equivariant as well~\cite{cohen2016group}. For segmentation tasks, the last layer of a scale-equivariant network is a pooling layer over the scale-dimension.

Finally, we propose scale-equivariant transposed convolutions (upconvolutions). Transposed convolutions have been shown to be equivariant in the group-convolution setting in general but have so far only been used in the rotation-equivariant setting\cite{winkens2018improved}. Transposed convolutions can be helpful in creating CNN architectures for medical image segmentation, such as the U-Net~\cite{cciccek20163d}.

\begin{figure}[tb]
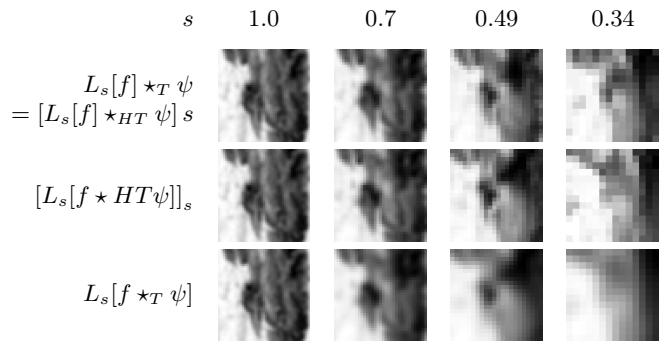

    \centering
    \include{figures/equivariance_vis}
    \caption{Comparison of 2D slices of the 4D/3D output of scale-equivariant and non-scale-equivariant convolution.
    Scaling $L_s$ of the input $f$ followed by scale-equivariant convolution $\star_{HT}$ with a filter $\psi$ (first row) yields almost the same result as scale-equivariant convolution followed by scaling and selection $[\cdot]_s$ of the respective ``scale slice'' along the scale dimension of the feature map on $HT$ (second row).
    This shows that $\star_{HT}$ is scale-equivariant apart from small interpolation artifacts.
    On the other hand, the ordinary (i.e.~non-scale-equivariant) convolution $\star_T$ followed by scaling (third row) yields a different result, and thus is not scale-equivariant.
    Note that due to the three-dimensional nature of the data, the 2D~slices are not just scaled versions of each other but also strongly influenced by values of neighbouring slices when scaled, depending on the scale.}
    \label{fig:exp:equivariance}
\end{figure}

\section{Experiments}
\subsection{Experimental Setup}
We perform brain tumor segmentation on the BraTS~2020 dataset. The inputs are four MRI contrasts (T1w, T1w with gadolinium, T2w, and T2w-FLAIR) with image registration applied.
The output targets are voxel-wise annotations of three different tumor classes (Gd-enhancing tumor, peritumoral edema, and necrotic and non-enhancing tumor core) obtained through annotations of up to four raters and approved by neuroradiologists~\cite{bakas2017advancing, bakas2018identifying, menze2014multimodal}.
In our experiments, we restricted the output target to binary labels representing healthy tissue or tumors.
As we did not have access to the validation and test set of the BraTS challenge, our full dataset consists of 369 samples of which we used up to 250 for the training of our models.
We applied instance normalization to the samples for feature scaling, which performed slightly better than other feature scaling methods.
Since training-data augmentation by random scaling has been shown\cite{sosnovik2019scale} to be beneficial also for scale-equivariant methods (possibly due to introducing intermediate scales not present in the discretized group, and increasing training-data diversity due to interpolation artifacts), we study the effect of scaling training samples with a random scale between~$0.7$ and~$1.0$ in every training step in additional dedicated experiments.

We base our model on the U-Net architecture~\cite{cciccek20163d} using four downsampling and upsampling blocks (with 4, 8, 16, and 32 channels from top to bottom). Down- and upsampling are performed using strided and transposed scale-equivariant convolutional layers, respectively (kernel size~$5$, stride~$2$), each followed by two scale-equivariant convolutional layers, and residual connections are used within blocks. Skip-connections are used between blocks of the same resolution except for the uppermost layer. The truncated scaling group used in the scale-equivariant network is $\{0.9^0, 0.9^1, 0.9^2, 0.9^3\} = \{1.0, 0.9, 0.81, 0.729\}$ and the kernel basis consists of 27 basis functions, constructed as described in Section~\ref{sec:basis}.
Experiments were carried out to compare max- vs. avg-pooling over the scaling group in the last layer.

We compare our model to a baseline using the same architecture but with 16, 32, 64 and 128 channels in the respective network blocks. 
The network width and depth, learning rate and other hyperparameters were tuned using manual search over a wide range separately for the baseline method and the scale-equivariant methods. Training was performed for up to 160 epochs, depending on the size of the training set, using an NVIDIA RTX 8000 GPU, up to 5 GB VRAM and 10 GB RAM. The training lasted about 2 hours on average.

The loss used is a sum of the soft Dice loss~\cite{milletari2016v} and the binary cross-entropy, and the Adam optimizer with a learning rate of 0.01 was used for training with an exponential learning rate decay.

\begin{figure}[tb]
    \centering
    \subfigure{
\centering
\begin{tikzpicture}
\begin{axis}[
    width= 0.4\textwidth,
    height=0.7\textwidth,
    xlabel={Number of training samples},
    ylabel={Average Dice coefficient on test data},
    xmin=8, xmax=27,
    ymin=0.13, ymax=0.67,
    xtick={10, 15, 25},
    ytick={0.2, 0.3, 0.4, 0.5, 0.6},
    legend style={at={(0.79,0.02)},anchor=south},
    legend cell align={left},
    ymajorgrids=true,
    ylabel near ticks,
    grid style=dashed,
    log ticks with fixed point,
]

\addplot[
    color=red,
    dashed,
    mark=square,
    mark options=solid
    ]
    coordinates {
    (10, 0.18882323738751042)(15, 0.3874996092553963)(25, 0.5977172465316027)
    };
\addplot[
    color=red,
    mark=square,
    ]
    coordinates {
    (10, 0.1565232962733839)(15, 0.3218923866337932)(25, 0.5672396593040419)
    };
\addplot[
    color=blue,
    dashed,
    mark=triangle,
    mark options=solid
    ]
    coordinates {
    (10, 0.5218366138177275)(15, 0.46850029235922913)(25, 0.5979143452636912)
    };
\addplot[
    color=blue,
    mark=triangle,
    ]
    coordinates {
    (10, 0.43551647015465744)(15, 0.45982772040546216)(25, 0.6399402090230909)
    };
\addplot[
    color=plotgreen,
    dashed,
    mark=o,
    mark options=solid,
    ]
    coordinates {
    (10, 0.6049321221391193)(15, 0.5360411418955466)(25, 0.6281835572661034)
    };
\addplot[
    color=plotgreen,
    mark=o,
    ]
    coordinates {
    (10, 0.607525467218346)(15, 0.6071835835744478)(25, 0.6445059039794346)
    };    
\end{axis}
\end{tikzpicture}
}
\subfigure{
\centering
\begin{tikzpicture}
\begin{axis}[
    width= 0.58\textwidth,
    height= 0.472\textwidth,
    xlabel={Scaling of test data},
    ylabel={Dice coefficient on test data},
    xmin=0.65, xmax=1.05,
    ymin=0.81, ymax=0.89,
    xtick={0.7, 0.8, 0.9, 1.0},
    ytick={0.82, 0.84, 0.86, 0.88},
    legend style={at={(0.4,-0.9)},anchor=south, draw=none},
    legend cell align={left},
    ymajorgrids=true,
    grid style=dashed,
    ylabel near ticks,
    x label style={at={(axis description cs:0.5,-0.08)},anchor=north},
]
\addplot[
    color=plotgreen,
    dashed,
    mark=o,
    mark options=solid,
    ]
    coordinates {
    (0.7,0.839)(0.8,0.867)(0.9,0.865)(1.0,0.882)
    };
\addplot[
    color=plotgreen,
    mark=o,
    ]
    coordinates {
    (0.7,0.854)(0.8,0.861)(0.9,0.864)(1.0,0.867)
    };
\addplot[
    color=blue,
    dashed,
    mark=triangle,
    mark options=solid
    ]
    coordinates {
    (0.7,0.833)(0.8,0.854)(0.9,0.856)(1.0,0.863)
    };
\addplot[
    color=blue,
    mark=triangle,
    ]
    coordinates {
    (0.7,0.868)(0.8,0.877)(0.9,0.875)(1.0,0.876)
    };
\addplot[
    color=red,
    dashed,
    mark=square,
    mark options=solid
    ]
    coordinates {
    (0.7,0.817)(0.8,0.836)(0.9,0.836)(1.0,0.846)
    };
\addplot[
    color=red,
    mark=square,
    ]
    coordinates {
    (0.7,0.840)(0.8,0.848)(0.9,0.854)(1.0,0.846)
    };

    \legend{
    \footnotesize{Ours (avg-pooling)},
    \footnotesize{Ours (avg-pooling) + augmentation},
    \footnotesize{Ours (max-pooling)},
    \footnotesize{Ours (max-pooling) + augmentation},
    \footnotesize{Baseline},
    \footnotesize{Baseline + augmentation},
    }
\end{axis}
\end{tikzpicture}
}
    \caption{Comparison of methods in terms of quality of brain tumor segmentation.
    (Left:)
    Performance comparison when trained on less training data.
    The Dice scores are averaged over all test data scalings $\{0.7,0.8,0.9,1.0\}$; results are similar for each individual scaling factor.
    Scale-equivariant neural networks considerably outperform the baseline method when trained on only few samples.
    (Right:) Generalization of trained methods to scaled test data. The scale-equivariant models consistently outperform the baseline method.}
    \label{fig:dice-plots}
\end{figure}
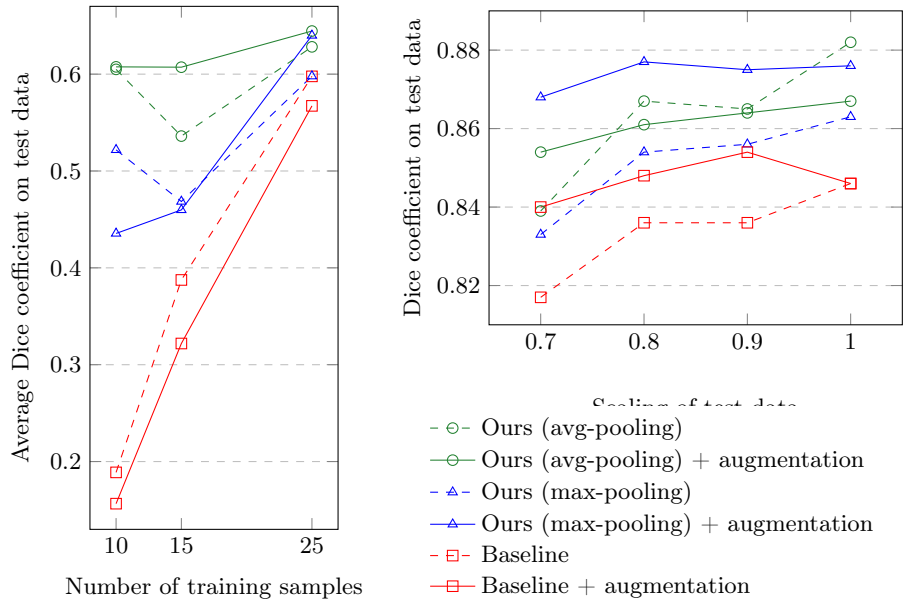

\subsection{Results and Discussion}

\begin{figure}[htb!]
    \centering
    \input{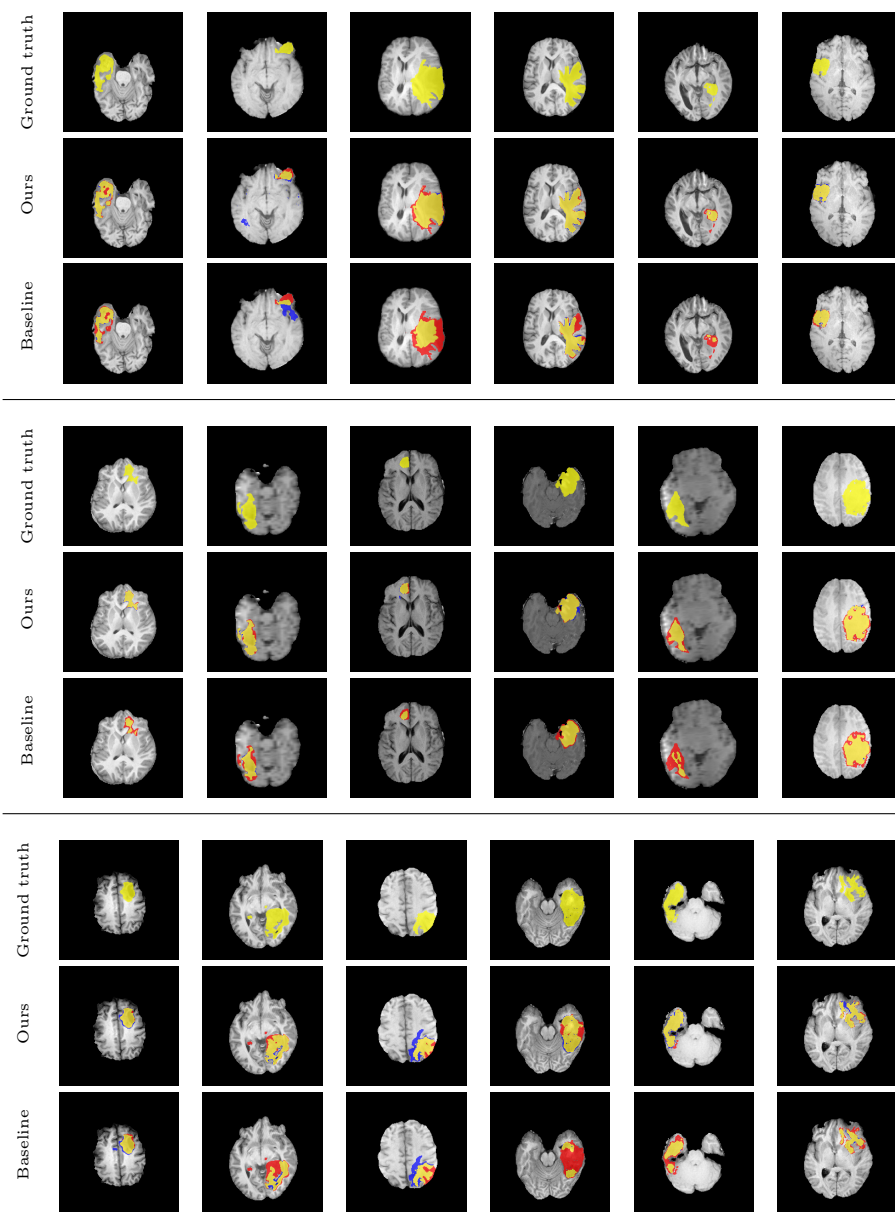}
    \caption{Visualization of the ground truth segmentation and the predictions of the scale-equivariant method and the baseline method. True positives are shown in yellow, false positives in blue, and false negatives in red. Our proposed method generally performs better even with complex lesion shapes.}
    \label{fig:brain-examples}
\end{figure}

Our proposed method benefits from its inherent scale-equivariance which is robust against scale changes as can be seen in Fig.~\ref{fig:exp:equivariance}. It outperformed the baseline on all scalings of the test data (see Fig.~\ref{fig:dice-plots}).
The scale-equivariant network using average-pooling and no data augmentation reached a Dice score of $0.882 \pm 0.104$ on the (non-scaled) test set, and thus outperforms the method using max-pooling ($0.876 \pm 0.110$) and the non-scale-equivariant baseline method ($0.846 \pm 0.128$) that are using training-data augmentation (see Tab.~\ref{tab:test-augmented}).

\begin{table}[tb]
\caption{Quality of brain tumor segmentation using the scale-equivariant neural network (ours) and baseline neural network, both trained with data augmentation. The scale-equivariant neural network generalizes to a broad range of scales and outperforms the baseline on all scales of test data.
}
\label{tab:test-augmented}
\centering
\begin{adjustbox}{width=\textwidth}
\begin{tabular}{lcccccccc}
\hline
Scaling of test data                             & \multicolumn{2}{c}{1.0}           & \multicolumn{2}{c}{0.9}           & \multicolumn{2}{c}{0.8}           & \multicolumn{2}{c}{0.7} \\ \hline
\multicolumn{1}{l|}{Method}           & \begin{tabular}[c]{@{}c@{}}Ours\end{tabular} & \multicolumn{1}{c|}{Baseline} & \begin{tabular}[c]{@{}c@{}}Ours\end{tabular} & \multicolumn{1}{c|}{Baseline} & \begin{tabular}[c]{@{}c@{}}Ours\end{tabular} & \multicolumn{1}{c|}{Baseline} & \begin{tabular}[c]{@{}c@{}}Ours\end{tabular} & Baseline \\\hline
\multicolumn{1}{l|}{Dice coefficient} & $\bf0.876$ & \multicolumn{1}{c|}{$0.846$} & $\bf0.875$ & \multicolumn{1}{c|}{$0.854$} & $\bf0.877$ & \multicolumn{1}{c|}{$0.848$} & $\bf0.868$ & $0.840$ \\
\multicolumn{1}{l|}{} & $\bf\pm 0.110$ & \multicolumn{1}{c|}{$\pm 0.128$} & $\bf\pm 0.115$ & \multicolumn{1}{c|}{$\pm 0.131$} & $\bf\pm 0.099$ & \multicolumn{1}{c|}{$\pm 0.140$} & $\bf\pm 0.103$ & $\pm 0.154$ \\
\hline
\multicolumn{1}{l|}{Balanced accuracy} & $\bf0.940$ & \multicolumn{1}{c|}{$0.916$} & $\bf0.944$ & \multicolumn{1}{c|}{$0.923$} & $\bf0.933$ & \multicolumn{1}{c|}{$0.920$} & $\bf0.920$ & $0.919$ \\
\multicolumn{1}{l|}{} & $\bf\pm 0.055$ & \multicolumn{1}{c|}{$\pm 0.078$} & $\bf\pm 0.053$ & \multicolumn{1}{c|}{$\pm 0.072$} & $\bf\pm 0.060$ & \multicolumn{1}{c|}{$\pm 0.078$} & $\bf\pm 0.070$ & $\pm 0.085$ \\
\hline
\end{tabular}
\end{adjustbox}
\end{table}

Training on an augmented dataset increased the performance of the baseline model in dealing with scaled versions of the test data. Data augmentation proved to be beneficial for the scale-equivariant method as well, as it stabilizes training and improves the networks' capability to deal with interpolation artifacts introduced through artificial scaling of the data. This is evidenced by the better performance of the models on scaled test data (see Fig.~\ref{fig:dice-plots}) and is consistent with observations for scale-equivariant methods in the two-dimensional domain~\cite{sosnovik2019scale}. Our method trained with data augmentation and using max-pooling outperforms the baseline trained with and without data augmentation on all scalings of the test data. Visualizations of the segmentation results are shown in Fig.~\ref{fig:brain-examples}.

In a separate experiment, we evaluate the data efficiency of the proposed method. The scale-equivariant method was less affected by a reduction in training data than the baseline method: for a small number of training data with only 10 or 15 training samples, the proposed method performed up to three times better than the baseline method. Thus, it is a valuable tool in the medical setting, where training data is often limited due to high data-acquisition costs, privacy, or rare diseases.

\section{Conclusions}
Our method shows strong results and consistently outperforms the baseline. Max- and average-pooling yield slightly different results depending on the context.
Average-pooling uses information from various scales simultaneously and thus can use ``fractal (multi-scale) properties''~\cite{weibel1991fractal} of the image. Due to relying on several scales at once, these image properties get easily destroyed by interpolation artifacts, or move beyond the truncation boundary of the scale dimension (see Section~\ref{sec:truncated-group}), when scaling the training data or test data. This might explain why average-pooling trained without data augmentation outperforms all other methods on unscaled test data, but is more negatively affected by scaling of test data. On the other hand, max-pooling is not affected at all by artifacts at the truncation boundary if they have smaller magnitude than the maximal values selected by max-pooling.

We propose a range of scale-equivariant neural network layers that can be used to analyse medical images, but can also be applied in other fields with 3D voxelized data. The formulation of scale-equivariant networks could further be extended to point cloud data or other 3D data representations.
Our proposed method outperformed the non-scale-equivariant baseline and showed its efficiency in a low-resource setting, which can be especially helpful in the medical area.
\bibliographystyle{splncs04}
\bibliography{bibliography}
\end{document}